\documentclass[runningheads]{llncs}
\usepackage{graphicx}
\usepackage{hyperref}
\begin{document}
\title{Unmasking the Uniqueness: A Glimpse into Age-Invariant Face Recognition of Indigenous African Faces}

\titlerunning{Unmasking the Uniqueness: ...}

\author{Fakunle Ajewole\inst{1},
Joseph Damilola Akinyemi\inst{2},
Khadijat Tope Ladoja\inst{1},
Olufade Falade Williams Onifade\inst{1}}
\authorrunning{Ajewole et al.}

\institute{University of Ibadan, Ibadan, Nigeria\\
\email{fajewole5054@stu.ui.edu.ng}\\
\email{\{kt.bamigbade,ofw.onifade\}@ui.edu.ng}\\
\and
University of York, Heslington, United Kingdom\\
\email{joseph.akinyemi@york.ac.uk}}
\maketitle            
\begin{abstract}

The task of recognizing the age-separated faces of an individual, Age-Invariant Face Recognition (AIFR), has received considerable research efforts in Europe, America, and Asia, compared to Africa. Thus, AIFR research efforts have often under-represented/misrepresented the African ethnicity with non-indigenous Africans. This work developed an AIFR system for indigenous African faces to reduce the misrepresentation of African ethnicity in facial image analysis research. We adopted a pre-trained deep learning model (VGGFace) for AIFR on a dataset of 5,000 indigenous African faces (FAGE\_v2) collected for this study. FAGE\_v2 was curated via Internet image searches of 500 individuals evenly distributed across 10 African countries. VGGFace was trained on FAGE\_v2 to obtain the best accuracy of 81.80\%. We also performed experiments on an African-American subset of the CACD dataset and obtained the best accuracy of 91.5\%. The results show a significant difference in the recognition accuracies of indigenous versus non-indigenous Africans.


\end{abstract}
\section{Introduction}
\label{sec:introduction}
"Biometrics" refers to the computerized identification of individuals based on genetic and behavioural characteristics. With the increasing use of biometrics, identification, and verification are now more dependable and accurate, proving each person's true identity. We have a variety of biometric tools at our disposal, including hand geometry, voice, fingerprint, iris, and card recognition. Face recognition provides several advantages over the other biometrics stated above due to its high accuracy, non-contact nature, user-friendliness, speed, and dependability ~\cite{Sun2021}

In many applications, including access control, distribution of government benefits, and criminal investigations, age-invariant facial recognition is crucial as subjects age with time. A robust identification system should be able to successfully identify people despite age differences between the enrolled and the probe images of the same individual. Face recognition that is age-invariant may also aid in reducing operational expenditures by minimizing the necessity for re-enrollment ~\cite{Zhao2022}.

Despite several research efforts in Age-Invariant Face Recognition (AIFR), there is little or no representation of African ethnicity in the results of published research~\cite{DamilolaAkinyemi2023}. In the few cases of representation of African ethnicity, it has been wrongly represented as African-American in MORPH ~\cite{Ricanek2006}, and CACD ~\cite{Chen2014}. In ~\cite{Akinyemi2021}, it was shown that there are significant differences in the ageing patterns of indigenous Africans and non-indigenous Africans. Indigenous Africans refer to Africans who reside within the continent and are therefore affected by the climate, economy, and conditions of living within the African continent, while non-indigenous Africans refer to Africans who reside outside the African continent (e.g., African-Americans).

This work addresses a critical and significant issue in the field of face recognition: the misclassification and inappropriate grouping of Indigenous Africans with continental Africans within existing age-invariant face recognition systems. This challenge has resulted in denials of identity and inaccurate recognition for individuals from the Indigenous African population~\cite{Akinyemi2021}. The difficulty is significant because human subjects' appearances in training or enrollment photographs can differ significantly from the appearance presented for identification. Motivated by the work of \cite{Akinyemi2021} on the FAGE dataset (first used in \cite{onifade2014gw}) which contained images representing only one African country, we curated a new dataset cutting across individuals from 10 African countries evenly distributed within the 5 geographical zones of the continents. We also extracted an African-American subset of a large AIFR dataset, CACD~\cite{Chen2014}, for comparison with the curated dataset. Our findings include:

1.	The first dataset of indigenous African faces with a wide coverage of the African continent and a balanced number of age-separated face images per subject.

2.	A promising and competitive recognition accuracy on the curated dataset.

3.	Significant differences in the performance on the curated dataset when compared with the performance on the African-American counterpart under the same experimental settings.

The next section \ref{sec:litreview} presents a review of the various approaches that have been used for face recognition with machine learning. Section \ref{sec:method} describes the methodology (Materials and Methods) Section \ref{sec:result} describes the model performance results, discussions, and improvements from our findings, and Section \ref{sec:conclusion} concludes the paper.

\section{Literature Review}
\label{sec:litreview}
This section reviews the research that has been conducted over the years on age-invariant face recognition. Researchers throughout the world have employed several ML techniques for age-invariant face recognition, and a discussion of their methods, datasets, and results is presented in this section.

The authors of ~\cite{Park2010}, proposed a 3D ageing modelling technique to address the challenge of achieving temporal invariance in automatic face recognition. They used three different face databases, FG-NET ~\cite{Panis2016}, MORPH ~\cite{Ricanek2006}, and BROWNS to evaluate the effectiveness of their approach. The proposed method resulted in substantial performance improvements on all three databases, which demonstrates the effectiveness of the proposed ageing modelling method.

In ~\cite{Gong2015}, an innovative face representation and matching was presented for the age-invariant face recognition challenge. A unique maximum entropy feature descriptor (MEFD) technique was developed that encodes the microstructure of face photos into a sequence of discrete codes. The coding entropy is improved to obtain discriminative and descriptive information from densely sampled encoded face pictures. An identity factor analysis technique was created to estimate the possibility that two given faces had the same core identity. The technique was assessed on the FGNET and MORPH datasets, and accuracies of 76.2\% and 92.26\% respectively, were reported.

According to ~\cite{Chen2016}, deep convolutional neural networks (DCNN) have achieved top results on numerous tasks in computer vision, including face verification. DCNN models can not only classify big data fluctuations but also learn a compact and discriminative feature representation when the amount of training data is sufficiently high. The proposed approach by ~\cite{Wang2018}, decomposes deep facial characteristics into two orthogonal elements that depict age-related and identity-related aspects. This led to identity-related traits that are resilient to ageing for age-invariant face recognition. Specifically, the deep facial features were decomposed in the spherical coordinate system comprising the radial coordinate $r$ and the angular coordinate $\phi$.

The proposed Application Invariant Model (AIM) in ~\cite{Zhao2022}, addressed the challenges in recognizing faces across ages by providing a unified deep architecture that could learn powerful age-invariant facial representations and perform attention-based face rejuvenation and ageing, which could be used for age-invariant face recognition and other applications.

In ~\cite{Akinyemi2021}, the authors observed that most face age estimation works had explored the influence of continental ethnicities (e.g., African, Asian, Caucasian, and Hispanic) and that there were yet to be critical experimentations on the impact of local ethnicities such as indigenous Africans. Extending previous studies such as \cite{Akinyemi2016,onifade2015gwageer}, they highlighted that the ageing of African Americans may not be the same as that of Africans who reside on the African continent and referred to the ethnicity of the latter group as 'indigenous' Africans. Using subsets of the CACD and FAGE, they investigated these supposed differences in age estimation and reported that they indeed existed. However, the dataset subsets used were limited in size, which questioned their results' validity.

The individualized face pairing method developed by ~\cite{DamilolaAkinyemi2023}, matches faces versus full groups of faces arranged by individuals. Similarity score vectors were created for both matching and non-matching image-individual pairings, and the vectors were employed for age-invariant face recognition. This technique used the individualized aspects of ageing to lessen the influence of ageing on face recognition, enabling individuals to be accurately identified across their multiple age-separated face photos. Their model achieved high recognition accuracies on FAGE, FG-NET and CACD, but did not investigate the ethnic variations in the datasets.

According to ~\cite{Dhamija2022}, the suggested technique to boost the performance of AIFR incorporates an updated ASM architecture to extract handmade and deep features for AIFR, in combination with a 7-layer CNN architecture and a reduced picture size of 32x32 pixels to decrease delay time and space complexity. The results show that the suggested method beats state-of-the-art algorithms in face recognition and achieves good accuracy throughout the age spectrum. The presented methodology achieves a maximum accuracy of 91.76\% for the LAG database ~\cite{Bianco2017}, outperforming all existing state-of-the-art methodologies.

In ~\cite{Hou2022}, authors proposed a novel age-invariant face recognition framework using Mutual Information Minimization (MT-MIM), which disentangles the mixture of face features into two nearly independent components: the identity-dependent component and the age-dependent component. They introduced a multi-task learning framework based on mutual information minimization to partition face representations into identity-dependent and age-dependent components for age-invariant face recognition. They evaluated MT-MIM on popular public-domain face ageing datasets FG-NET ~\cite{Panis2016}, MORPH  Album 2~\cite{Ricanek2006}, CACD ~\cite{Chen2014}, and AgeDB ~\cite{Moschoglou2017}, and obtained significant improvements over previous state-of-the-art methods. Specifically, their method exceeds the baseline models by over 0.4\% on MORPH Album 2, and over 0.7\% on CACD subsets, which are impressive improvements over the high accuracy levels of above 99\% and an average of 94\%.

~\cite{Zhang2023}, proposed a method for age-invariant face recognition based on identity-age shared features (ISF). Specifically, they decoupled facial representation into three parts, i.e., pure identity features, pure age features, and identity-age shared features, to improve the independence of age and identity features, thereby reducing interference from age-related information and improving the accuracy of face recognition. They conducted experiments on several benchmark datasets to evaluate the performance of the proposed method. Experimental results on benchmark datasets for face ageing (FG-NET 95.67\%, AGE-DB 97.53\%, CALFW 96.03\%, and CACD-VS 99.58\%) show that the proposed ISF outperforms state-of-the-art AIFR approaches. Overall, the authors provided a new perspective for face recognition and demonstrated the potential of their proposed method to be applied in various fields, such as security, surveillance, and human-computer interaction.
In ~\cite{Tripathi2022}, the authors presented a local feature computation method to extract local features from all pixels within pre-defined local regions of the face. Their method involved a four-step procedure which includes the computation of Local Difference Pattern (LDP) and Local. Directional Gradient Relation Pattern (LDGRP) which are then used to create a histogram-based facial feature vector. Their experiments demonstrated 90.75\% and 96.95\% recognition accuracy on the FG-NET and MORPH datasets, respectively. It's interesting to see that this method achieves such a significant performance despite the use of local features and without any deep learning methods.

MTLFace ~\cite{Huang2023}, is a recent effort to perform the tasks of AIFR and face synthesis jointly. It involved the collection of a large dataset, which is aimed at becoming a new benchmark for the two tasks. Extensive experiments and comparisons were conducted, especially across the age dimension, specifically focusing on young versus old faces. However, there is no mention of the consideration of ethnic representation or diversity in the dataset, as far as we know. The dataset indeed contains face pictures of subjects across various ethnicities, but the indigenous nature of these faces remains in question.

The results presented in the article   ~\cite{Zhang2023} demonstrate that the proposed method, which includes introducing identity-age shared features and utilizing a two-stage constraint algorithm, outperforms state-of-the-art approaches in age-invariant face recognition. The experimental evaluation conducted on benchmark datasets (FG-NET, AGE-DB30, CALFW, and CACD-VS) shows superior performance compared to existing methods such as DAL and MTLFace. The study highlights the effectiveness of the proposed approach in improving the accuracy and robustness of cross-age identity recognition in face recognition systems.

From the above-reviewed works, AIFR systems seem to have improved over the years, but not enough attention seems to have been given to the indigenous African ethnicity. This presents a significant drawback to face recognition given the population of the African continent, and it is to this end that this work investigates AIFR on indigenous Africans and compares it with AIFR on non-indigenous Africans.

\section{Methodology}
\label{sec:method}
Given the success of existing AIFR research on large datasets, we have observed that most datasets have limited representations of Africans. In most cases where Africans are included in large AIFR datasets (e.g., MORPH ~\cite{Ricanek2006}, and CACD ~\cite{Chen2014}, these Africans were mostly non-resident within the African continent. As subtle as this sounds, research has shown that it has a significant impact on the results of facial image analysis ~\cite{Guo2010}, ~\cite{Akinyemi2021}, This has motivated us to conduct AIFR on indigenous African faces (i.e., Africans living on the African continent) vis-à-vis non-indigenous Africans (i.e., Africans living outside the African continent).

In our methodology, we formulated the AIFR problem as a multiclass closed-set recognition problem to determine how well we can recognize the individuals within the dataset. To achieve this, we set out formulations for the AIFR problem as follows:

Given a set, A, of face images organized by individuals as 

\begin{equation}
  A = \{B_i | i = 1,2, …, n \}
  \label{eq:eq1}
\end{equation}

\begin{equation}
  B = \{b_k | k = 1,2, …, m \}
  \label{eq:eq2}
\end{equation}

Where $b_k$ is a single-face image of an individual and $m$ is the number of age-separated images of individual $i$. The order of images in $A$ and $B$ are unimportant, hence the use of sets in equations \ref{eq:eq1} and \ref{eq:eq2}.

Based on the definitions in Equations \ref{eq:eq1} and \ref{eq:eq2}, we aim to find a model that appropriately maps $B$ to $A$ as defined in \ref{eq:eq3}. So, the function $g(b_k)$ should return the appropriate $B_i \in A$ to which $b_k$ belongs.

\begin{equation}
g(b_k): B \rightarrow{} A	
  \label{eq:eq3}
\end{equation}

\begin{figure}[t]
  \centering
   \includegraphics[width=0.8\linewidth]{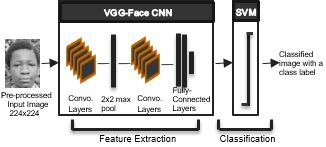}

   \caption{A deep learning-based AIFR model using VGG-FACE}
   \label{fig1}
\end{figure}

To realize the function $g()$, we employed a deep learning method to learn facial features and classify faces based on input facial images. Then, we experimented with two datasets, one collected for this research and the other, an existing dataset. Figure ~\ref{fig1} shows the overall design of our methodology where an input face is preprocessed and passed through a deep learning architecture (VGGFace) ~\cite{Cao2018}, to extract relevant identity features and classify (recognize) the facial image. Our motivation for using the VGGFace architecture is that it is already trained to extract facial features which makes it more reliable for extracting facial features for age-invariant face recognition.

\begin{figure}[t]
  \centering
   \includegraphics[width=0.8\linewidth]{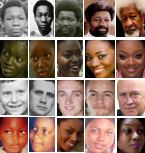}

   \caption{Sample of the faces in the FAGE\_v2 dataset. Each row shows the age-separated images of an individual.}
   \label{fig2}
\end{figure}

It is obvious that set $A$ in Equation \ref{eq:eq1} is the dataset of face images, which is organized by the individuals/subjects in the dataset. In our study, we have used two datasets; FAGE\_v2 (Facial expression, Age, Gender, and Ethnicity, version 2), a dataset of indigenous African faces collected for this research, and CACD, a publicly available face dataset. FAGE\_v2, a different set from the FAGE dataset in ~\cite{DamilolaAkinyemi2023}, was curated for this study by searching for and downloading images of 500 Africans across 10 African countries evenly distributed within the 5 geographical zones of the continent (i.e., North, East, South, West and Central Africa). Specifically, we identified 100 popular individuals from each zone to make a total of 500 individuals and collected 10 age-separated face images of each individual, summing up to a total of 5,000 face images. Samples of faces in the FAGE\_v2 dataset are shown in Figure ~\ref{fig2} and the dataset is available on \href{https://www.kaggle.com/datasets/ajewoleolaitan/fage-dataset}{Kaggle}. The individuals whose images were downloaded include politicians, clergy, athletes, successful business people, etc.

CACD contains images of actors of different ethnicities. We created a list of actors who identified as African-American or Black from imdb.com and used this list to extract the images of these individuals in CACD and we realized 8,656 images of 89 individuals. The details of both datasets are presented in Table \ref{tab1}. Upon further cleaning (removing duplicates and wrongly labeled images), we had 6979 images left.

Thus, $n$, the number of individuals, as in Equation \ref{eq:eq1}, is 500 for FAGE\_v2 and 89 for the CACD subset. The disparity in the number of subsets is huge, but it is also an indicator of the limited representation of African faces in the CACD dataset. While $k$, the number of images for each individual is fixed at 10 in FAGE\_v2, it is variable in CACD with an average of 82 images per subject. Still, many of those images are noisy and had to be removed as stated in the previous paragraph.

\begin{table}
\caption{Analysis of FAGE\_v2 and CACD.}
 \label{tab1}
\begin{center}
\begin{tabular}{|l|l|l|}
\hline
Parameters &  FAGE v2 & CACD\\
\hline
Total Number of Images &  {5,000} & 163,446\\
\hline
Total Number of Individuals &  {500} & 2000\\
\hline
Num. of Indigenous African Individuals & {500} & 0\\
\hline
Num. of African-American individuals & {0} & 88\\
\hline
Num. of African-American images & {0} & 8,656\\
\hline
Age range  & {1-95} & 20-60\\
\hline
Average num. of images per individual & {10} & 82\\
\hline
Average age range of each individual  & {43} & 40\\
\hline
Num. of images used for comparative analysis & {880} & 880\\
\hline
\end{tabular}
\end{center}
\end{table}

\subsection{The AIFR Model for Indigenous African Faces}
We adopted the VGGFace ~\cite{Cao2018}, VGGFace was trained on VGGFaxe2 dataset, Each input image was automatically preprocessed by detecting them with MTCNN ~\cite{Zhang2019}, and cropping the face regions, and resizing the same to 224 x 224, the input dimensions for the VGGFace architecture. VGGFace comprises 13 convolutional layers for extracting facial features and 3 fully connected layers, the last of which is the classification layer. We removed the classification layer and replaced it with varying numbers of fully connected layers including dropout layers as indicated in Tables ~\ref{tab2} and ~\ref{tab3}. In all the experiments, we used the Adam optimizer with a learning rate of 0.0001 and the categorical cross-entropy loss. The optimization algorithm, Adam, was used to update the network's weights based on the training data. The model was fine-tuned by adjusting hyper-parameters such as the Epoch, learning rate, batch size, etc. We explored different combinations of hyperparameter settings to train and validate the model across both datasets and compared the results.

Each dataset was randomly split into training, validation, and test sets. For FAGE\_v2, there were 3500, 500, and 1,000 images in the training, validation, and test splits, respectively. For CACD, there were 4,491 images for training, 1,081 images for validation, and 1,307 images for testing. The predictions of the trained model were evaluated using classification accuracy, precision, recall, and F1-score all of which are represented as percentages. Accuracy measures the overall correctness of predictions, precision gauges the proportion of true positives in positive predictions, recall assesses the proportion of true positives captured, and the F1-score is the harmonic mean of precision and recall.

\section{Experiments, Results, and Discussions}
\label{sec:result}

For each dataset, we conducted a set of 5 experiments using different hyperparameter combinations and reported these details in ~\ref{tab2} and ~\ref{tab3}. The five settings, as seen in ~\ref{tab2} and ~\ref{tab3} involve different values of the training iterations (epochs), batch size, number of dense layers, and drop-out fraction. In all cases, the results are reported on the test set of each dataset. The dense layer, as in ~\ref{tab2} and ~\ref{tab3}, indicates the number of fully connected layers added to the architecture to replace the original classification layer.

For both datasets, the best accuracies was obtained at 50 training epochs, a batch size of 64, one fully connected layer, and 50\% dropout layer. In this experimental setting, we obtained the best accuracy of 81.8\% for FAGE\_v2 and 91.5\% for CACD. In terms of the number of fully connected layers, the results seem to improve with a smaller number of them. This indicates a good representation of the main VGGFace architecture, and only a few final layers may be needed to have a good classification. The higher drop-out rate of 0.5 (50\%) always produced better results, except when the batch size was too large (128). Generally, 50 training epochs were sufficient to achieve good performance, and a higher number of epochs did not seem to improve the results so much.

\begin{table}
\caption{Recognition accuracies on FAGE v2 dataset}
\label{tab2}
\begin{center}
\begin{tabular}{|l|l|l|l|l|l|}
\hline
Setting & Epoch & Batch size & Dense & Drop out & Accuracy (\%)\\
\hline
1 & 50 & 64 & 1 & 0.5 & 81.8 \\
\hline
2 & 64 & 16 & 3 & 0.5 & 81.6 \\
\hline
3 & 50 & 32 & 1 & 0.2 & 81.1 \\
\hline
4 & 50 & 128 & 1 & 0.5 & 77.5 \\
\hline
5 & 64 & 16 & 4 & 0.2 & 77.5 \\
\hline

\end{tabular}
\end{center}
\end{table}

\begin{table}
\caption{Recognition accuracies on CACD dataset}
\label{tab3}
\begin{center}
\begin{tabular}{|l|l|l|l|l|l|}
\hline
Setting & Epoch & Batch size & Dense & Drop out & Accuracy (\%)\\
\hline
1 & 50 & 64 & 1 & 0.5 & 91.5 \\
\hline
2 & 64 & 16 & 3 & 0.5 & 91.0 \\
\hline
3 & 50 & 32 & 1 & 0.2 & 90.0 \\
\hline
4 & 50 & 128 & 1 & 0.5 & 87.5 \\
\hline
5 & 64 & 16 & 4 & 0.2 & 88.5 \\
\hline

\end{tabular}
\end{center}
\end{table}

Apart from the accuracy, we evaluated AIFR performance on the test sets of both datasets using precision, recall, and F1-score, and the results are presented in Tables ~\ref{tab4} and ~\ref{tab5}. The results presented in Tables ~\ref{tab2} and ~\ref{tab3} show that 81.8\% and 91.5\% are the best-performing accuracies of the 5 experimental settings, which is setting 1 (see Tables ~\ref{tab2} and ~\ref{tab3}). As seen from Tables ~\ref{tab4} and ~\ref{tab5}, the accuracy, precision, recall, and F1-score values are close for both datasets, indicating a fairly balanced recognition rate.

\begin{table}
\caption{Performance evaluation of FAGE\_v2}
\label{tab4}
\begin{center}
\begin{tabular}{|l|l|l|l|l|l|}
\hline
Metrics & Output (\%)\\
\hline
  Accuracy & 81.8 \\
\hline
  Precision & 83.4 \\
\hline
 Recall & 82.3\\
\hline
 F1-score & 82.8\\
\hline
\end{tabular}
\end{center}
\end{table}

\begin{table}
\caption{Performance evaluation of CACD}
\label{tab5}
\begin{center}
\begin{tabular}{|l|l|l|l|l|l|}
\hline
Metrics & Output (\%)\\
\hline
  Accuracy & 91.5 \\
\hline
  Precision & 93.1 \\
\hline
 Recall & 92.1\\
\hline
 F1-score & 92.6\\
\hline
\end{tabular}
\end{center}
\end{table}

\subsection{Comparative Analysis}
Due to the wide differences in the distribution of both datasets, we performed an additional experiment to provide a fair baseline for the comparison of their performances and to further verify the earlier experiments. In this comparative analysis, we selected 10 images from each of the 88 African-American individuals in CACD and also selected 88 individuals from FAGE so that both subsets had 880 images each. Due to the very small number, these subsets were each divided into training, validation, and test splits (70:10:20), and VGGFace was again trained for 50 epochs on each dataset, validated, and tested on the respective splits. The results revealed a similar pattern: 90.4\%, 79.7\%, and 83.5\% accuracies for training, validation, and test, respectively, on the CACD subset, and 83.5\%, 68.8\%, and 72.0\% accuracies for training, validation, and test, respectively, on the FAGE subset.

There are two things we would like to point out from the results. First, one can observe that in all cases, the CACD images were always better predicted than the FAGE\_v2 images. This is an indication of the differences in the ageing features of indigenous versus non-indigenous Africans, and this is consistent with the findings in ~\cite{Akinyemi2021}. This means the model can recognize non-indigenous African faces (in CACD) much better than the indigenous ones (in FAGE\_v2). This could also be largely due to the extent of diversity of the ethnicities on which the VGGFace architecture was pre-trained, as well as an indication of the misrepresentation or underrepresentation of the indigenous African ethnicity on the face database on which VGGFace was trained. Secondly, despite the disparities observed above, the results on FAGE\_v2 are quite promising, yet they indicate that more work needs to be done to push the results further up. Despite the relatively small number of images per individual in FAGE\_v2 (10) compared to CACD (82), we consider a $\approx10\%$ difference in accuracy rates on both datasets a good indication of the diversity of the FAGE\_v2 dataset and an indication that with only a little more effort, we can improve the task of recognizing indigenous Africans.

\section{Conclusion}
\label{sec:conclusion}
This work has investigated the differences in the ageing patterns of indigenous and non-indigenous Africans in the realm of Age-Invariant Face recognition. An AIFR model was developed for recognizing the age-separated faces of indigenous Africans to a reasonable degree of recognition accuracy (81.8\%). More importantly, this work uncovered the differences in the ageing patterns of indigenous and non-indigenous Africans using a subset of the CACD dataset. This creates room for further exploration of these differences for the development of robust face recognition algorithms for African ethnicities. Unfortunately, there are not enough individuals in the CACD dataset to match up with the number of individuals in FAGE\_v2, and reducing the number of individuals in FAGE\_v2 will only reduce the dataset to a minimal amount, not even able to match up with the size of the CACD subset. In future work, we hope to investigate a larger subset of non-indigenous Africans, probably harvested from different datasets other than CACD to combat this limitation. We also hope to keep enlarging the FAGE\_v2 dataset for further analysis.

%
%
\bibliographystyle{splncs04}
\bibliography{Myrefs}
%





\end{document}